\documentclass[10pt,twocolumn,letterpaper]{article}

\usepackage[pagenumbers]{cvpr}      

\definecolor{cvprblue}{rgb}{0.21,0.49,0.74}
\usepackage[pagebackref,breaklinks,colorlinks,allcolors=cvprblue]{hyperref}

\usepackage{multirow} 
\usepackage{ulem}
\usepackage{graphicx}

\title{InfoScale: Unleashing Training-free Variable-scaled Image Generation via Effective Utilization of Information}

 \author{Guohui Zhang$^{1}$\thanks{Equal contribution.} \quad Jiangtong Tan$^{1}$\footnotemark[1] \quad Linjiang Huang$^{2}$ \quad Zhonghang Yuan$^{1}$ \\ 
 \quad Mingde Yao$^{3}$ \quad Jie Huang$^{4}$ \quad Feng Zhao$^{1}$\thanks{Corresponding author}\\
 $^{1}$USTC $^{2}$Beihang University $^{3}$CUHK MMLab $^{4}$Kuaishou Technology\\
 }
\begin{document}

\let\oldtwocolumn\twocolumn
\renewcommand\twocolumn[1][]{
    \oldtwocolumn[{#1}{
    \vspace{-3em} 
    \begin{center}
    \includegraphics[width=\textwidth]{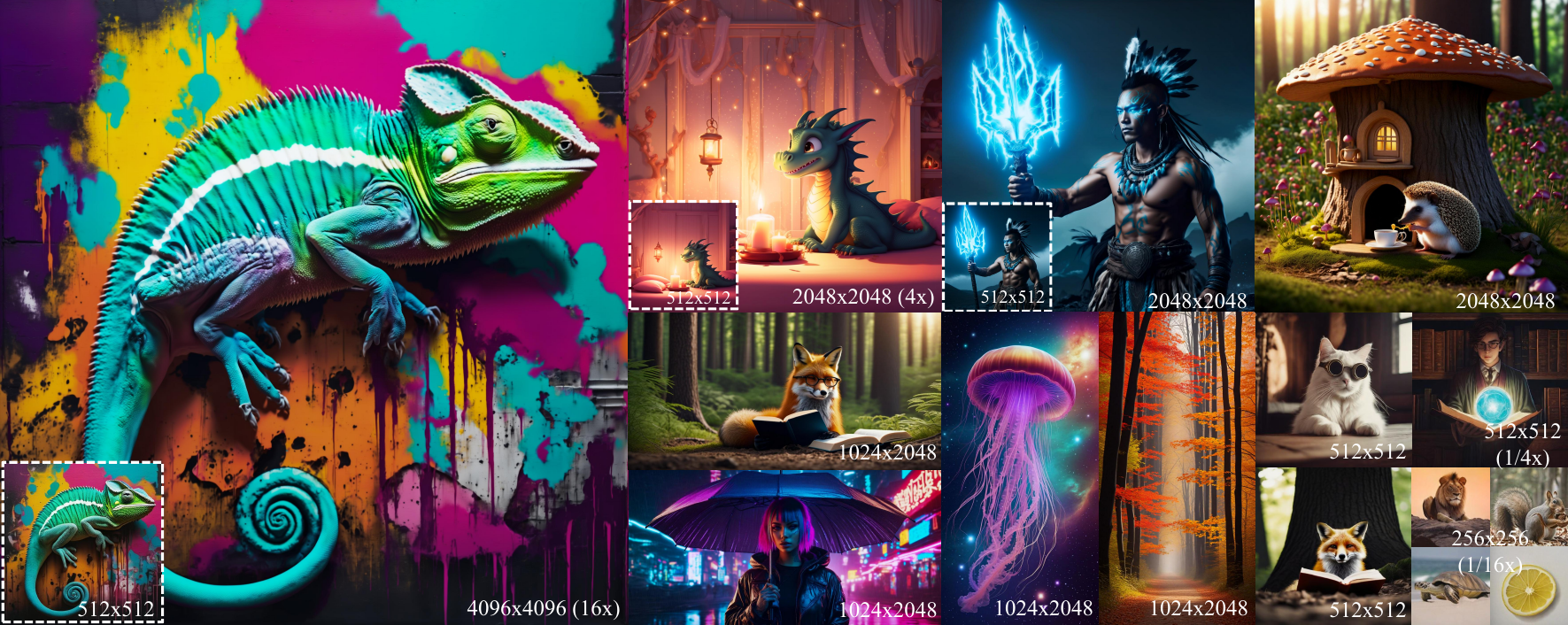}
    \setlength{\abovecaptionskip}{-0.1cm}
    \setlength{\belowcaptionskip}{-0.1cm}
    \captionof{figure}{Generated images of InfoScale based on SDXL from lower resolution to higher resolution. Our method extends SDXL to generate images from $\mathbf{1/16 \times, 1/4 \times}$ to $ \mathbf{4 \times, 16 \times}$ without any fine-tuning. Best viewed ZOOMED-IN.}
    \label{fig:mainapp}
    \end{center}
    }]
}

\maketitle
\begin{abstract}
Diffusion models (DMs) have become dominant in visual generation but suffer performance drop when tested on resolutions that differ from the training scale, whether lower or higher.
Current training-free methods for DMs have shown promising results, primarily focusing on higher-resolution generation. However, most methods lack a unified analytical perspective for variable-scale generation, leading to suboptimal results.
In fact, the key challenge in generating variable-scale images lies in the differing amounts of information across resolutions, which requires information conversion procedures to be varied for generating variable-scaled images.
In this paper, we investigate the issues of three critical aspects in DMs for a unified analysis in variable-scaled generation: dilated convolution, attention mechanisms, and initial noise.
Specifically, 1) dilated convolution in DMs for the higher-resolution generation loses high-frequency information. 
2) Attention for variable-scaled image generation struggles to adjust the information aggregation adaptively. 
3) The spatial distribution of information in the initial noise is misaligned with variable-scaled image. 
To solve the above problems, we propose \textbf{InfoScale}, an information-centric framework for variable-scaled image generation by effectively utilizing information from three aspects correspondingly. 
For information loss in 1), we introduce Progressive Frequency Compensation module to compensate for high-frequency information lost by dilated convolution in higher-resolution generation.
For information aggregation inflexibility in 2), we introduce Adaptive Information Aggregation module to adaptively aggregate information in lower-resolution generation and achieve an effective balance between local and global information in higher-resolution generation.
For information distribution misalignment in 3), we design Noise Adaptation module to re-distribute information in initial noise for variable-scaled generation. 
Our method is plug-and-play for DMs and extensive experiments demonstrate the effectiveness in variable-scaled image generation. Code is \href{https://github.com/zghhui/InfoScale}{available}

\end{abstract}    
\section{Introduction}
\label{sec:intro}
Diffusion models (DMs) have witnessed remarkable progress in visual generation~\cite{rombach2022high,ho2020denoising,song2020denoising}, which converts the information from initial noises to image space. 
Despite the powerful generation capabilities of DMs, for variable-scaled image generation that is often required in practical applications, directly inputting initial noise with resolutions lower or higher than the training resolution setting usually leads to visual defects, such as incomplete content in lower-resolution images and distorted structure in higher-resolution images.
This is a challenging issue since the information conversion procedures are different in variable-scaled image generation, and the components in DMs (i.e., convolution, attentions) are over-optimized to process the information in training settings.
Although fine-tuning DMs is a choice, it requires substantial computation resources and high-quality data.

Recently, quantities of training-free approaches for variable-scaled image generation have emerged and have attracted widespread attention.
\textbf{Primarily},  most of them are focused on higher-resolution generation~\cite{he2023scalecrafter,huang2024fouriscale,du2024demofusion,qiu2024freescale,zhang2023hidiffusion}.
One line of training-free-based higher-resolution generation methods primarily relies on incorporating dilated convolution or samplings (i.e., ScaleCrafter~\cite{he2023scalecrafter} and FouriScale~\cite{huang2024fouriscale}) to align higher-resolution image structure information with the training resolution.
While another line focuses on firstly generating the main structure in the training resolution and refining the details in the higher resolution (i.e., DemoFusion~\cite{du2024demofusion} and FreeScale~\cite{qiu2024freescale} ), which also requires dilated convolution or samplings in higher-resolution generation procedures to avoid artifacts appearing.
However, the dilated convolution used in these methods often leads to details information losses in higher-resolution generation.
\textbf{Meanwhile}, Only a few works~\cite{jin2023training,haji2024elasticdiffusion} focus on variable-scaled (\textbf{both lower and higher resolution}) image generation. However, their generated results often lack sufficient detail or require large latency overheads.

\begin{figure}[htbp]
\centering
\includegraphics[width=0.45 \textwidth]{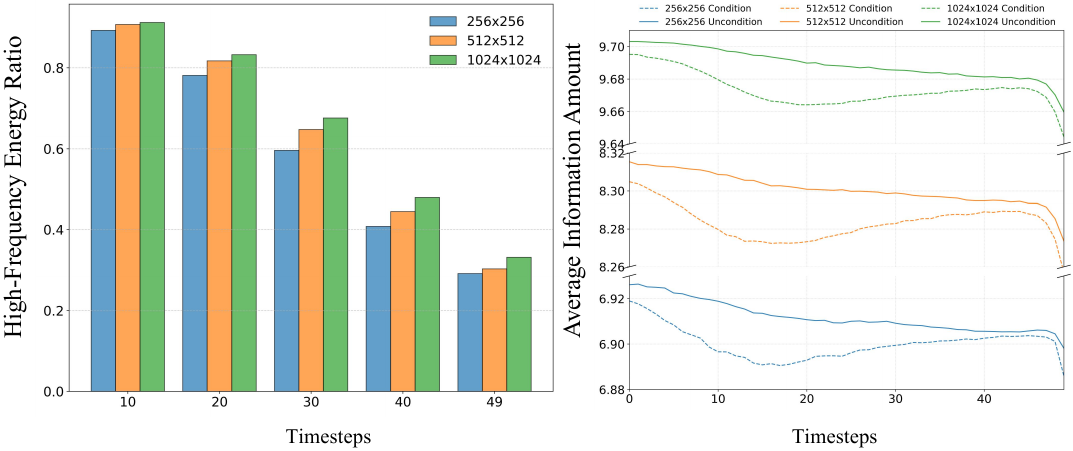}
\\
  \vspace{-0.1in}
\caption{The left and right figures illustrate that higher-resolution image contain greater proportion of high-frequency components and larger information amount.}
\label{fig:high_freq_diff_res}
  \vspace{-0.2in}
\end{figure}

Although the aforementioned approaches have made great efforts, a unified analytical perspective for variable-scaled (both lower and higher resolution) generation has rarely been discussed.
In fact, the key challenge in generating variable-scaled images is that the information amount of the generated image is varied across resolutions, as shown in Fig.~\ref{fig:high_freq_diff_res}. 
Higher-resolution images or latents generally contain a greater amount of information and larger proportions of high-frequency components, while lower-resolution ones behave oppositely. 
Since DMs are only optimized to convert the initial noise to the generated image in the level of training-resolution information amount, the contradiction in information conversion procedures across resolutions constrains the potential of applying DMs in generating variable-scaled images.

In this paper, we investigate the problems of three critical aspects in DMs for a unified analysis in variable-scaled generation: dilated convolution, attention mechanisms, and initial noise. Specifically,
\textbf{1)} In higher-resolution generation, directly applying dilated convolution loses some high-frequency information, as shown in Fig.~\ref{fig: dialte_conv_9_10_19_20}, which prevents generating more image details.
\textbf{2)} In lower-resolution generation, scaled attention struggles to aggregate enough information in the limited contextual range. 
However, in higher-resolution generation, attention tends to aggregate redundant and repetitive information, as shown in Fig.~\ref{fig:low_high_attn}. This inflexible information aggregation leads to unreasonable information utilization. 
\textbf{3)} In lower-resolution generation, DMs struggle to handle non-uniform information distribution (i.e., lower entropy) in initial noise compared to training setting's initial noise, resulting in incomplete content. In higher-resolution generation, the information distribution of initial noise is over-uniform and contains multiple responses to the prompt, causing the information in these regions to be processed independently into repetitive objects, see Fig.~\ref{fig:NA_noise_4}.

Therefore, in order to achieve effective utilization of information without the aforementioned three information utilization obstacles, we propose \textbf{InfoScale}, an information-centric variable-scaled image generation framework, including three key designs corresponding to them respectively. 
Specifically, for the information loss in \textbf{1)}, we introduce Progressive Frequency Compensation (PFC) module to extract high-frequency components from cached noise of the previous timestep to compensate for the predicted noise at the current timestep when applying dilated convolution in higher-resolution generation. 
For information aggregation inflexibility in \textbf{2)}, we introduce Adaptive Information Aggregation (AIA) module to adaptively adjust information aggregation ability of attention. We design dual-scaled attention (DSAttn) based on the original scaled attention by adjusting the attention entropy to be more adaptive, enhancing information aggregation ability of attention in lower-resolution generation. We further fuse the features of DSAttn and original attention to effectively balance local-enhanced information (aggregated by DSAttn) and global information (original Attn) in higher-resolution generation. 
For information distribution misalignment in \textbf{3)}: we introduce the Noise Adaptation (NA) module which enhances the uniformity of information distribution in the central region to encourage information aggregation in lower-resolution generation. We gradually suppress the uniformity of information distribution from centric to surrounding region through the NA module to alleviate the repeated response to the prompt in higher-resolution generation.
The designs of our method are training-free and are flexibly plug-and-play for DMs. Extensive experiments demonstrate that our framework significantly improves the visual quality in variable-scaled image generation.

Our core contributions can be summarized as follows:
\begin{itemize}
    \item We propose \textbf{InfoScale}, an information-centric variable-scaled image generation framework, offering a unified analytical perspective for variable-scaled image generation.
    \item We design progressive frequency compensation, adaptive information aggregation and noise adaptation modules to achieve efficient information utilization.
    \item Extensive experiments validate the effectiveness of our framework by plugging into DMs in a training-free way.
\end{itemize}
\section{Related Work}
\label{sec:related work}
\subsection{Text-to-image generation}
Diffusion models (DMs)~\cite{dhariwal2021diffusion,li2024hunyuan,liu2024llm4gen,zhuo2024lumina} have attracted widespread attention due to their excellent image generation quality. Denoising Diffusion Probabilistic Models (DDPM)~\cite{ho2020denoising} demonstrated the potential of DMs in image generation. Moreover, Classifier-Free Guidance (CFG)~\cite{ho2022classifier} ennobled DMs to generate images confirming to given prompts. Since operations in pixel space require substantial computational resources, Latent Diffusion Models (LDM)~\cite{rombach2022high} proposed to transfer the diffusion process to latent space~\cite{blattmann2023align,he2022latent}, thereby reducing the training burden and laying the foundation for high-resolution image generation. Thanks to large-scale training data~\cite{schuhmann2022laion}, the Stable Diffusion series~\cite{podell2023sdxl,rombach2022high} has achieved groundbreaking progress in visual generation.

\subsection{Variable-scaled image generation}

\begin{figure}[hbtp]
\centering
\includegraphics[width=0.45 \textwidth]{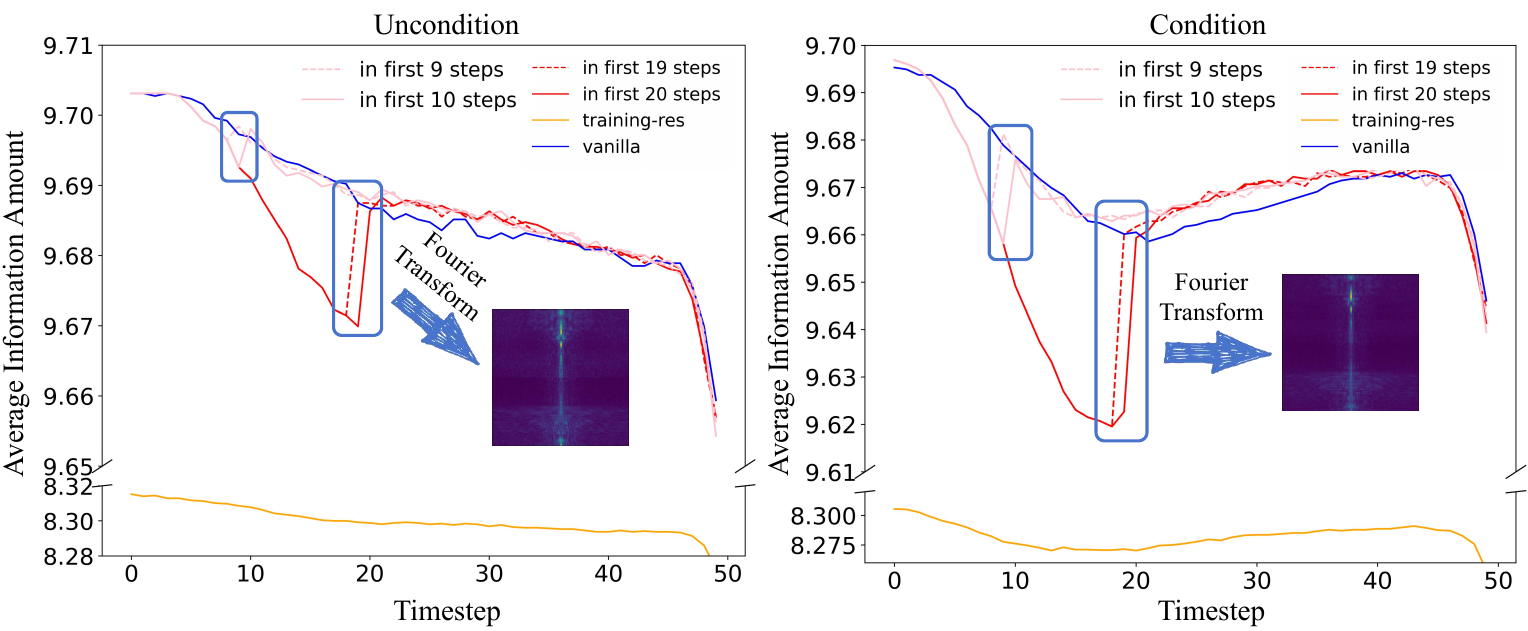}
\\
\caption{Information loss in dilated convolution. It can be observed that during the steps using dilated convolution, the information amount shows significant decrease, indicating that dilated convolution reduces redundant information, while frequency analysis shows that these information includes some high-frequency components. Vanilla refers to no dilated convolution.}
\label{fig: dialte_conv_9_10_19_20}
\end{figure}

Due to trained on limited resolutions, directly applying pre-trained diffusion models to generate images with novel resolutions often results in visual defects, such as incomplete content at lower-resolution images and repeated objects or distorted structures at higher-resolution images. For higher-resolution generation, some approaches propose training or fine-tuning models with higher-resolution images to improve performance of models~\cite{hoogeboom2023simple,liu2024linfusion,ren2024ultrapixel,chen2024pixart}.
However, the scarcity of high-resolution image data and the significant increase in computational resource demands due to resolution scaling limit the applicability of such methods. Many training-free approaches propose using specific strategies during inference to fully leverage the potential of diffusion models in higher-resolution image generation~\cite{hwang2024upsample,lee2023syncdiffusion,kim2024diffusehigh,lin2024cutdiffusion,yang2024fam,zhang2024frecas,lu2023hiprompt,yang2025rectifiedhr}. ScaleCrafter~\cite{he2023scalecrafter} and FouriScale~\cite{huang2024fouriscale} achieve structural consistency across different resolution by incorporating dilated convolution, while HiDiffusion~\cite{zhang2023hidiffusion} dynamically resizes features to align with the training resolution. Nevertheless, these methods still suffer from degraded image details.
MultiDiffusion~\cite{bar2023multidiffusion} extends to larger resolutions by generating overlapping patches. DemoFusion~\cite{du2024demofusion}, AccDiffusion~\cite{lin2024accdiffusion}, and FreeScale~\cite{qiu2024freescale} all first generate images at the training resolution to provide guidance for higher-resolution generation, yet their requires dilated convolution or samplings in higher-resolution generation procedures to avoid artifacts appearing. challenging. For variable-scaled image generation including lower-resolution generation, Attn-SF~\cite{jin2023training} adjusts attention entropy to achieve variable-scaled image generation, which  has much space for improvement.
\section{Method Motivation and Discussion}

\begin{figure}
\centering
\includegraphics[width=0.48 \textwidth ]{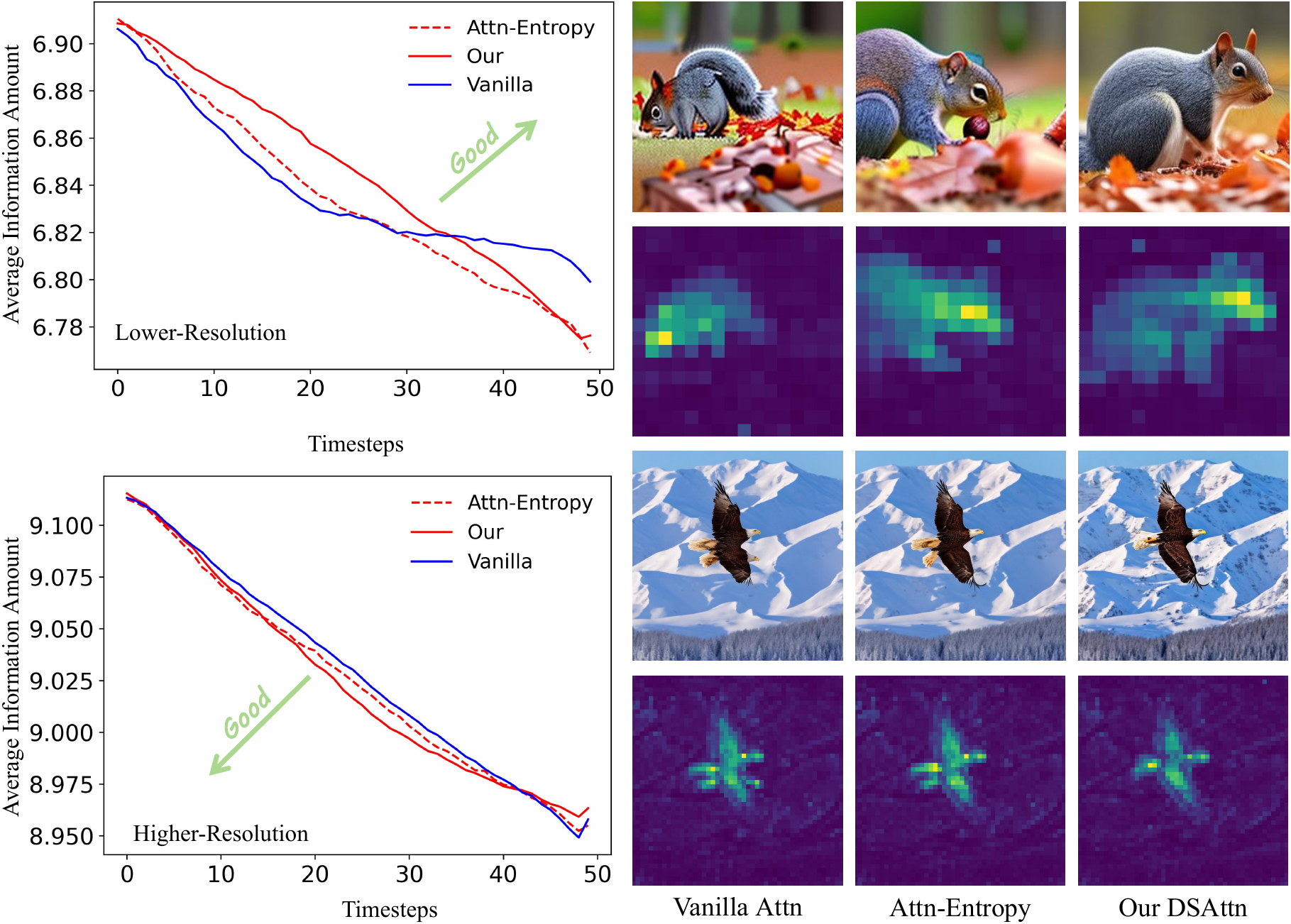}
\\
\caption{The top and bottom figures illustrate the inflexible aggregation ability of model for lower- and higher-resolution generation, respectively. We use dual scaled factors to achieve wider aggregation for lower-resolution generation, and vice versa for higher-resolution generation.}
\label{fig:low_high_attn}
\end{figure}

\subsection{Information Loss}
\label{Information Loss}
Information entropy~\cite{shannon1948mathematical} is a fundamental concept in information theory. In this work, we calculate it based on the self-attention scores from DMs, which quantifies the information conversion procedures across different resolutions. The information entropy \(H(X)\) is defined as:
\begin{equation}
H\left(\boldsymbol{X}\right)=-\sum_{i=1}^np\left(x_i\right)\log p\left(x_i\right)
    \label{information}
\end{equation}
where \(X\) is attention score after Softmax operation and \(i\) represents each position.

Directly scaling pre-trained diffusion models to higher resolutions results in generating images with repetitive objects.
This is because the model needs to process more information for higher-resolution generation, as shown in Fig.~\ref{fig:high_freq_diff_res}. 
To mitigate this repetition issue, Scalecrafter~\cite{he2023scalecrafter} replaces the standard convolution in U-Net with dilated convolution with a large receptive field, which has proven to be effective and is commonly used by further higher-resolution generation works (i.e., FourierScale~\cite{huang2024fouriscale} and FreeScale~\cite{qiu2024freescale}), but it inevitably leads to image quality degradation with details losses. This raises the question of what causes this phenomenon. 

We conduct experiments on 100 randomly generated prompts in higher-resolution generation to investigate the impact of dilated convolutions. We analyze the changes of information entropy during the sampling under different configurations. As shown in Fig.~\ref{fig: dialte_conv_9_10_19_20}, compared to standard convolution, dilated convolution tends to align the information to the training resolution in both conditional and unconditional sampling. This suggests that dilated convolution effectively compresses redundant information in higher-resolution generation, contributing to generating correct structures. 

To further analyze the difference of information in predicted noise using dilated convolutions in the early stages, we performed frequency analysis for both unconditional and conditional predicted noise. The results show that, compared to standard convolutions, dilated convolutions cause the predicted noise to lose high-frequency information. 
In fact, high-frequency components generally occupy a larger proportion in high-resolution images compared to low-resolution images, as shown in Fig.~\ref{fig:high_freq_diff_res}. Therefore, the loss of this high-frequency information potentially harms the quality of the generated image. To take advantage of dilated convolution in reducing redundant information while mitigating the loss of high-frequency information, frequency compensation is required to address this critical information-utilization bottleneck, see Sec.~\ref{sec:PFC}.

\begin{figure}[htbp]
\centering
  \includegraphics[width=0.48 \textwidth ]{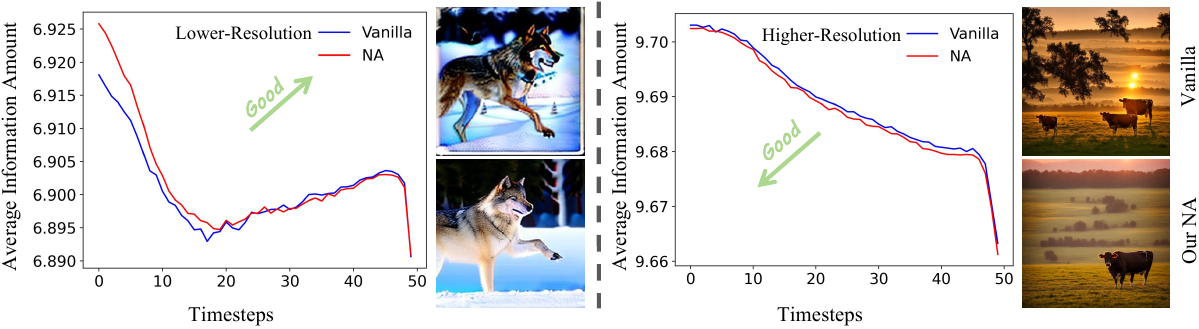}
  \\
  \caption{The left figure illustrates increasing the variance to adjust the information distribution of the initial noise promotes the information aggregation in lower resolution generation. The high-resolution generation in the right figure is the opposite.}
  \label{fig:NA_noise_4}
\end{figure}

\begin{figure*}[htb]
\centering
  \includegraphics[width=0.98\textwidth]{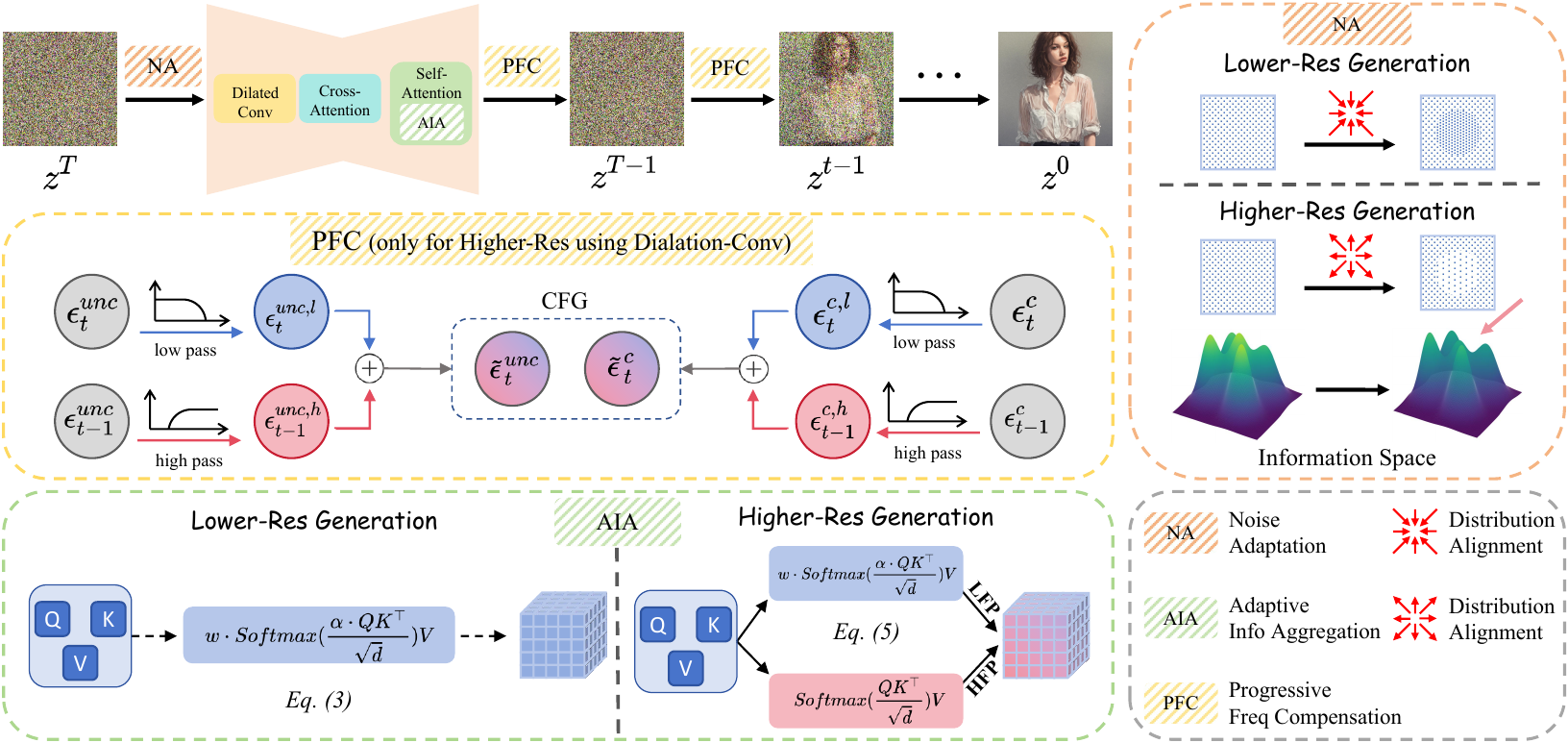}
  \\
  \caption{Overall framework of InfoScale. (a) In higher resolution generation, the Noise Adaptation (NA) module first modulates the initial noise according to resolution. Then, the Progressive Frequency Compensation (PFC) module extract high-frequency components from cached noise of the previous timestep to compensate for the predicted noise at the current timestep when applying dilated convolution. The Adaptive Information Aggregation module further fuse local (blue) and global information (red). (b) In lower-resolution generation, we also use NA module and replace original self-attention layer with DSAttn.}
  \label{fig:main}
\end{figure*}

\subsection{Information Aggregation Inflexibility}
\label{Information Aggregation Inflexibility}
For DMs, the scaled dot-product attention allows the model to focus on prominent parts of the input during the generation, facilitating efficient information aggregation.

We conduct experiments on information entropy and attention maps to analyze, see Fig.~\ref{fig:low_high_attn}. The results indicate that the self-attention focused solely on narrower local regions at lower-resolution images, failing to fully utilize global information to generate complete content. Meanwhile, the self-attention attempts to attend to more redundant and repetitive information in higher resolutions, leading to the generation of repeated structures. 
In comparison, Attn-SF~\cite{jin2023training} applies adaptive aggregation of information according to resolutions, but it is still limited to only scaling the value inside the Softmax. This approach leaves room for further improvements in scaling self-attention, as it fails to fully exploit the potential of adaptive aggregation that could dynamically adjust across different resolutions (see Sec.~\ref{sec:AIA}).

\subsection{Information Distribution Misalignment}
\label{Information Distribution Misalignment}
The distribution of initial noise is important for DMs. InitNo~\cite{guo2024initno} reveals that initial noise includes both semantically consistent and inconsistent components.
Inspired by this, we devote into the spatial distribution of information in initial noise. 
For higher resolution, the spatial distribution of initial noise is more uniform (i.e., more like Gaussian) than the training resolution, which contains multiple regions that respond to prompts, mainly manifested as multi-peak in spatial distribution. 
For lower resolution, the spatial distribution of initial noise is less uniform (i.e., less like Gaussian) than the training resolution, which struggles to form a completely effective region to respond to the prompts, resulting in incomplete content. 
We conducted experiments in Fig.~\ref{fig:NA_noise_4} that appropriately adjust the spatial distribution of the central area with variance scaling to align the training distribution. 
The experimental results show that this alignment improve information entropy, generating lower-resolution images with complete structures and higher-resolution image without repetitive objects. 
Therefore, we can further improve the utilization of information by adjusting a more appropriate initial noise distribution (see Sec.~\ref{sec:NA}).
\section{Method Implementation}
We propose \textbf{InfoScale}, which consists of three modules that correspond to the above three analyses respectively: Progressive Frequency Compensation (PFC) module, Adaptive Information Aggregation (AIA) module and Noise Adaptation (NA) module. 

\begin{table*}[htbp]
  \centering
  \caption{Quantitative comparison results. - indicates no data available for this metric. +Our is plug-and-play}
  \vspace{-0.1in}
   \resizebox{1\linewidth}{!}{
    \begin{tabular}{c|c|cccc|cccc|cccccc}
    \hline
    \multirow{2}{*}{Method} & \multirow{2}{*}{Scaling Factor} & \multicolumn{4}{c|}{\textbf{SD1.5}} & \multicolumn{4}{c|}{\textbf{SD2.1}} & \multicolumn{6}{c}{\textbf{SDXL}} \\
     \cline{3-16}
     & & \textbf{FID}$\downarrow$ & \textbf{KID}$\downarrow$ & \textbf{FID$_c$}$\downarrow$ & \textbf{KID$_c$}$\downarrow$
     & \textbf{FID}$\downarrow$ & \textbf{KID}$\downarrow$ & \textbf{FID$_c$}$\downarrow$ & \textbf{KID$_c$}$\downarrow$
     & \textbf{FID}$\downarrow$ & \textbf{KID}$\downarrow$ & \textbf{FID$_c$}$\downarrow$ & \textbf{KID$_c$}$\downarrow$ &
     \textbf{Clip}$\uparrow$ & \textbf{Time}$\downarrow$ \\

    \hline
     Direct-Inference & \multirow{4}{*}{0.5$\times$0.5} 
     & 114.76  & 0.031   & - & - 
     & 101.79  & 0.026   & - & - 
     & 96.31  & 0.017   & 78.72  & 0.021  
     &  30.26   &  \textcolor{red}{3s}\\
     
     Attn-SF~\cite{jin2023training} & & \uline{95.09}  & \uline{0.021}  & - & - 
     & \uline{88.34} & \uline{0.019}  & - & - 
     & 71.85  & 0.014  & \uline{49.20} & \uline{0.009} 
     &  30.62    &  \textcolor{red}{3s}\\

     ElasticDiffusion~\cite{haji2024elasticdiffusion} & & 94.40 & 0.030 & - & - 
     & 92.16 &  0.028  & - & - 
     & \textcolor{red}{70.45}  & \textcolor{red}{0.011} & 49.96 & \uline{0.009} 
     &  \uline{30.98}    &  \uline{50s}\\
     
     Ours & &  \textcolor{red}{90.34}  & \textcolor{red}{0.017} & - & - 
     & \textcolor{red}{82.93} & \textcolor{red}{0.016}   & - & - 
     & \uline{71.02} & \uline{0.013}  & \textcolor{red}{49.05}   & \textcolor{red}{0.008}  
     &  \textcolor{red}{31.04}    &  \textcolor{red}{3s}\\

    \hline
     Direct-Inf & \multirow{14}{*}{2$\times$2} & 86.97   & 0.014   &  46.45  &  0.009  
     & 80.82  & 0.011   &  40.56  &  0.007  
     &  117.07    &  0.033    &  128.05 & 0.041  
     &  31.45    &  35s\\
     
     Attn-SF~\cite{jin2023training} & & 82.28   &  0.011   & 45.45   & 0.007    
     &  79.81  &  0.010  & 37.87   & 0.006     
     & 111.86    &  0.030    & 124.17    & 0.035    
     &  31.55    &  35s\\
     
     HiDiffusion~\cite{zhang2023hidiffusion} & & 75.00   &  0.009   &  44.20  & 0.008  
     & 66.96   &  0.006  & 38.13   & 0.007   
     & 104.62    & 0.024      &  108.32   & 0.025   
     &  31.92    &  \uline{19s}\\
    
    MegaFusion~\cite{wu2025megafusion} & &  \uline{67.43} &  \uline{0.008} &  \uline{38.92} & \uline{0.007}  
     &  \uline{64.11} &  \uline{0.005} &  \uline{37.09} & \uline{0.007}
     &  72.38   & 0.007   & 93.06    & 0.018    
     &  32.47    &  \textcolor{red}{18s}\\
     
     DiffuseHigh~\cite{kim2024diffusehigh} & &  - &  - &  - & -  
     &  - &  - &  - & - 
     &  60.87   & 0.004    & 84.33    & 0.015    
     &  32.96    &  40s\\
     
     DemoFusion~\cite{du2024demofusion} & &  - &  - &  - & -  
     &  - &  - &  - & - 
     &  54.25   & 0.003    & 71.69    & 0.013    
     &  33.58    &  90s\\
     
     Accdiffusion~\cite{lin2024accdiffusion} & &  - &  - &  - & -  
     &  - &  - &  - & - 
     &  55.34   & 0.003    & 76.15   & 0.008    
      &  33.69    &  98s\\
     
     FreCaS~\cite{zhang2024frecas} & &  - &  - &  - & -  
     &  - &  - &  - & - 
     &  54.01   & 0.003    & 62.50    & 0.007   
      &  33.99    &  23s\\
     
     FouriScale~\cite{huang2024fouriscale} & &  68.81  & 0.008   & 39.79   & 0.007  
     &  65.22  &  0.006   & 38.19   &  0.007   
     &  78.17    & 0.017      & 93.75    & 0.025    
     &  32.22    &  65s\\
     
     FouriScale~\cite{huang2024fouriscale} +Our & & 68.16   & 0.008  &  39.03  & 0.007  
     &  63.74  &  0.005  &  36.84  & 0.006  
     & 77.47& 0.017  & 93.18 & 0.024 
     &  32.45    &  67s\\

     ScaleCrafter~\cite{he2023scalecrafter} & & 69.02  &  0.008  & 40.72   & 0.007   
     &  64.93  &  0.006  &  37.70  & 0.006    
     &  73.14    &  0.012     &  91.26   & 0.021    
     &  32.98    &  38s\\
     
     ScaleCrafter~\cite{he2023scalecrafter} +Our & & \textcolor{red}{66.34}   &  \textcolor{red}{0.007}  &  \textcolor{red}{37.97}  & \textcolor{red}{0.006} 
     & \textcolor{red}{62.73}    &  \textcolor{red}{0.004}  &  \textcolor{red}{36.47}  & \textcolor{red}{0.006}  
     & 72.57   & 0.012      & 91.03      &  0.021  
     &  33.08    &  40s\\     
     
     FreeScale~\cite{qiu2024freescale} & &  - &  - &  - & -
     &  - &  - &  - & - 
     &  \uline{51.99}      & \uline{0.003}    & \uline{60.99}    & \uline{0.006}  
     &  \uline{34.23}    &  47s\\
     
     FreeScale~\cite{qiu2024freescale} +Our & &  - &  - &  - & - 
     &  - &  - &  - & - 
     &   \textcolor{red}{50.79}      &  \textcolor{red}{0.002}    & \textcolor{red}{59.50}    & \textcolor{red}{0.004}   
     &  \textcolor{red}{34.26}    &  48s\\
    
    \hline
     Direct-Inf & \multirow{13}{*}{4$\times$4} & 180.47  & 0.062   & 58.56   & 0.020 
     &  173.75  &  0.058  & \uline{53.05}   & \uline{0.011}   
     &  189.08  & 0.078   & 165.43    & 0.059  
     &   30.17   &  504s\\
     
     Attn-SF~\cite{jin2023training} & &  169.05  &  0.054  &  57.72  & 0.018
     &  174.72  &  0.059  & \textcolor{red}{51.90}   & \textcolor{red}{0.009}   
     & 187.24   & 0.079   &  161.68   &  0.056  
     &  30.79    &  504s\\
     
     HiDiffusion~\cite{zhang2023hidiffusion} & & 135.00    &  0.043  & 62.66   & 0.027  
     & 119.76 &	0.033   &  76.03  &  0.026  
     & 144.08   & 0.056   &  186.45   & 0.079 
     &  31.34    &  \uline{138s}\\

     DiffuseHigh~\cite{kim2024diffusehigh} & &  - &  - &  - & -  
     &  - &  - &  - & - 
     &  75.43   & 0.022    & 115.40    & 0.031    
     &  32.07    &  557s\\
     
     DemoFusion~\cite{du2024demofusion} & &  - &  - &  - & -  
     &  - &  - &  - & -  
     & 60.60    & \textcolor{red}{0.006}     &  94.81    & 0.019  
     &  32.46    &  861s\\
     
     Accdiffusion~\cite{lin2024accdiffusion} & &  - &  - &  - & -  
     &  - &  - &  - & - 
     &  70.34   & 0.018    & 109.15    & 0.028    
     &  32.18    &  896s\\
     
     FreCaS~\cite{zhang2024frecas} & &  - &  - &  - & -  
     &  - &  - &  - & - 
     &  65.19   & 0.015    & 94.55    & 0.019    
     &  32.21    &  \textcolor{red}{130s}\\
     
     FouriScale~\cite{huang2024fouriscale} & & 76.63   & 0.011   & 57.19   &  0.019  
     & \uline{75.09}   & \textcolor{red}{0.009}   &  55.48  & 0.019  
     & 113.25  & 0.033   &  161.24  & 0.062  
     &  31.64    &  654s\\
     
     FouriScale~\cite{huang2024fouriscale} +Our & & 76.15   & 0.011   & 57.11   &  0.018  
     & \textcolor{red}{75.02}   & \textcolor{red}{0.009}   & 54.90   & 0.018   
     &  113.06   &  0.033   & 158.32   & 0.060   
     &  31.68    &  657s\\
     
     ScaleCrafter~\cite{he2023scalecrafter} & &  \uline{70.46}  &  \uline{0.008}  & \uline{53.41}   & \uline{0.016}  
     & 76.11   & 0.011   &  55.99  & 0.020   
     & 119.86    & 0.036   & 172.25      & 0.070 
     &  31.40    &  693s\\
     
     ScaleCrafter~\cite{he2023scalecrafter} +Our & & \textcolor{red}{70.37}   & \textcolor{red}{0.007}   & \textcolor{red}{52.52}   & \textcolor{red}{0.015} 
     &  75.47   & \uline{0.010}   &  55.69  & 0.020   
     &  119.10  &  0.035  &  170.98  & 0.068  
     &  31.47    &  698s\\
     
     FreeScale~\cite{qiu2024freescale} & &  - &  - &  - & - 
     &  - &  - &  - & - 
     &  \uline{60.16}   & 0.008    & \uline{94.20}    & \uline{0.019}    
     &  \uline{32.75}    &  532s\\
     
     FreeScale~\cite{qiu2024freescale}+Our & &  - &  - &  - & - 
     &  - &  - &  - & - 
     &  \textcolor{red}{59.74}     &  \uline{0.007}  &   \textcolor{red}{92.82}   &  \textcolor{red}{0.017}   
     &  \textcolor{red}{32.88}    &  534s\\
    \hline
    \end{tabular}
    }
  \label{tab:main}%
\end{table*}%

\subsection{Progressive Frequency Compensation} \label{sec:PFC}
Consider the hidden feature \( h \) and a convolution layer \( f \) that it will pass through, where the convolution kernel is \( k \). The dilated convolution operation \(\Phi_d(\cdot)\) can be represented as:
\begin{equation}
    f_{k}^d(h)=h\circledast\Phi_d(k),(h\circledast\Phi_d(k))(o)=\sum_{s+{d \cdot t}=p}h(p) \cdot k(q),
    \label{dilation_conv}
\end{equation}
where \(o, p, q\) are spatial locations used to index the feature and kernel, \(\circledast\) denotes convolution operation. This operation is equivalent to incorporating a down-sampling process before a up-sampling process, resulting in the loss of high-frequency information~\cite{huang2024fouriscale}. Here, we aim to to minimize the high-frequency information loss during the early stages of using dilated convolution.

We propose the Progressive Frequency Compensation module to address the loss of high-frequency information.
Considering the cumulative frequency loss in the iteration process of diffusion model and the continuity between consecutive latents, when predicting current noise $\epsilon_{t}$, we naturally employ the noise predicted in the previous step $\epsilon_{t-1}$ as the compensation for high-frequency information, which retains richer high-frequency information unaffected by dilated convolution. We identically process the conditional and unconditional noise. Denoting $N=\left\{unc, c\right\}$, it can be shown as:
\begin{equation}
\begin{aligned}
\epsilon_{t-1}^{N, h} & =\mathcal{FFT}(\epsilon_{t-1}^{N})\odot(1-\mathcal{H}), \\
\epsilon_{t}^{N, l} & =\mathcal{FFT}\left(\epsilon_t^{N}\right)\odot\mathcal{H}, \\
\tilde{\epsilon}_{t}^{N} & =\mathcal{IFFT}(\epsilon_{t-1}^{N, h}+\epsilon_{t}^{N, l}),
\end{aligned}
\label{e_fft}
\end{equation}
where \(\mathcal{FFT}\) is the Fast Fourier Transform and \(\mathcal{IFFT}\) is the Inverse Fast Fourier Transform. \(\mathcal{H}\) is the low-pass filter (LPF) with stop frequency of $D_0 = 0.25$ by default. 
More details can be find in supplementary material.

\begin{figure*}[htbp]
\centering
  \includegraphics[width=0.98\textwidth]{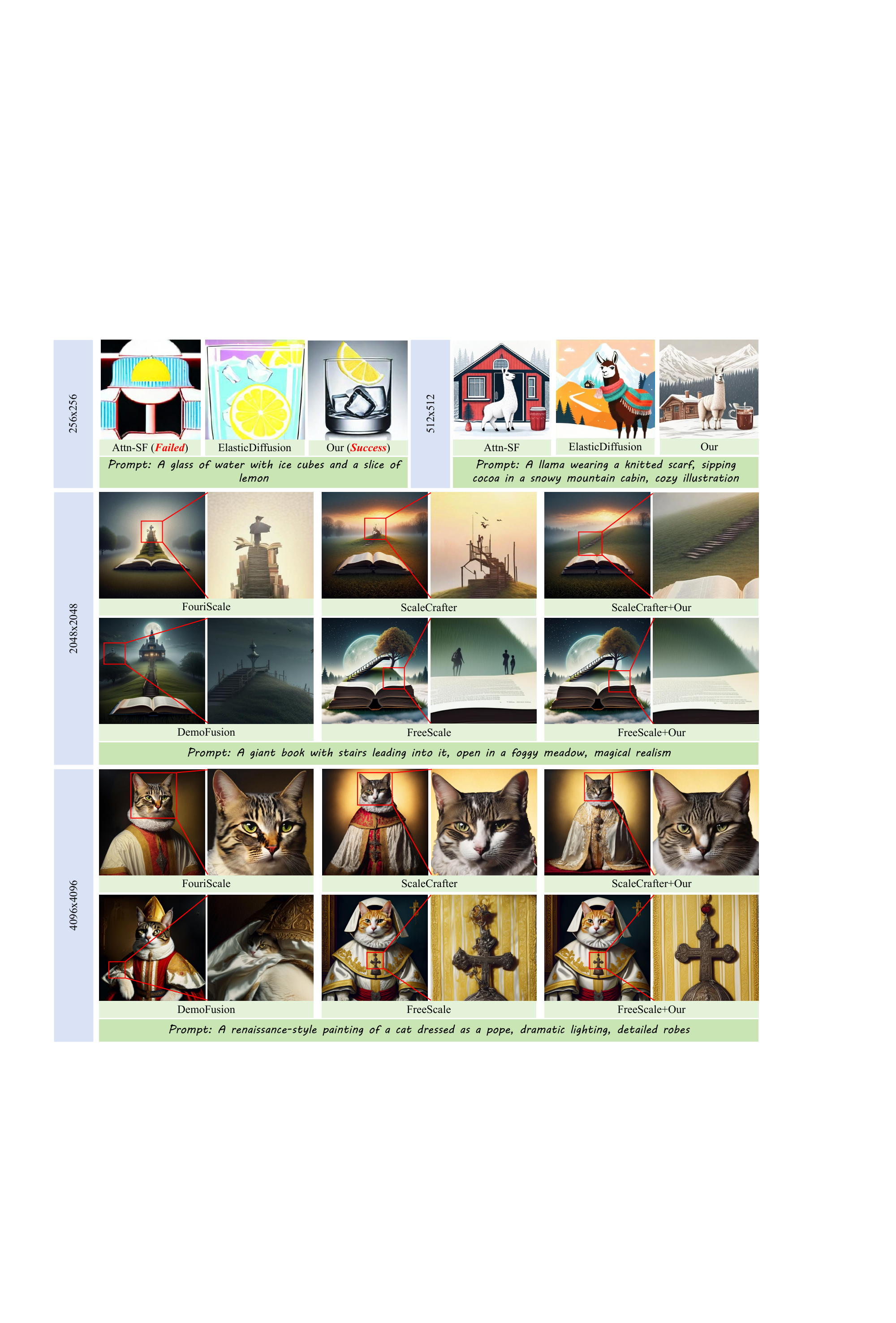}
  \\
  \caption{Qualitative comparison of images with other baselines. The training resolution of SDXL is $\mathbf{1024^2 (1/16\times)}$. Our method generates complete visual content on lower resolution, even up to $\mathbf{256^2 (1/16\times)}$. Best viewed ZOOMED-IN.}

  \label{fig: compare_baseline}
\end{figure*}

\subsection{Adaptive Information Aggregation} \label{sec:AIA}
In order to adaptively adjust the aggregation degree of information in self-attention, we propose Dual-Scaled Attention (DSAttn). Specifically, following prior work~\cite{jin2023training}, we introduce a scaled factor \(\alpha\) within the Softmax operation in self-attention layer, determined by the number of tokens in the current attention layer during both training and inference phases. Additionally, for low-resolution image generation, we incorporate an empirical hyperparameter $c_w$ to rescale the entire attention feature. This stems from our empirical observation that dual scaled factors allows attention to cover a broader range, thereby achieving better visual generation quality, which is shown below:
\begin{equation}
\begin{aligned}
\text{DSAttn(Q, K, V)} = w \cdot \text{Softmax}(\frac{\alpha \cdot QK^{^\top}}{\sqrt{d}})V,
\end{aligned}
\label{attn_scale}
\end{equation}
where $a$ and $w$ are scaled factors, $d$ is token dimension. Specifically, we calculate \(\alpha\) and \(w\) in the following form: 

\begin{equation}
\alpha = \sqrt{\log_L N}, \quad
w=
\begin{cases}
c_{w}& \text{L} > \text{N}\\
1& \text{L} \leq \text{N}
\end{cases}
\end{equation}
where $L, N$ are the number of tokens in self-attention in the training and testing phase. $c_w$ is hyperparameter with the value 0.75. In practice, we adopt different strategies for low-resolution and high-resolution generation.

\textbf{Lower-Resolution Generation.} We only replace the original attention layer with DSAttn in lower resolution generation to improve the ability to aggregate information.

\textbf{Higher Resolution Generation.} After using PFC module and DSAttn, the details of local features are effectively enhanced. To further leverage the original attention advantage in processing global information, we further fuse features from DSAttn and original attention to enhance global details and local structure. Specifically, we extract the low-frequency and high-frequency components of the features computed by DSAttn and the original attention layer, respectively. The low-frequency component is obtained by downsampling and then upsampling the feature, while the high-frequency component is obtained by subtracting the transformed result. Then, we fuse these two parts to obtain the enhanced feature.
\begin{equation}
\begin{aligned}
\tilde{h}_{t}= & \mathcal{U}\left(\mathcal{D}(h_t^s)\right) + \left(h_t-\mathcal{U}\left(\mathcal{D}(h_t)\right)\right),
\end{aligned}
    \label{h_fft}
\end{equation}
where $\mathcal{U}$ and $\mathcal{D}$ represent upsampling and downsampling operation (Nearest Neighbor interpolation) respectively.

\subsection{Noise Adaptation} \label{sec:NA}
We design a noise adaptation module to align the information distribution in the initial noise. Specifically, we modulate the initial noise ${Z}_T$ to obtain adaptive noise $\hat{Z}_T$ using a mask \(W\) with Gaussian weights. For different resolutions, W has different weights.

\begin{equation}
\begin{aligned}
\hat{Z}_T &= W \odot Z_T.
\end{aligned}
    \label{w_zt}
\end{equation}

\textbf{Lower-Resolution Generation.} The weight of \(W\) increases from the center to the surrounding to concentrate the information in the central area, which aims to generate complete object.

\textbf{Higher Resolution Generation.} The weight of \(W\) decreases from the center to the surrounding, which aims to mitigate distorted structure and repetitive objects.
\section{EXPERIMENTS}
\textbf{Experimental Settings.} To demonstrate the effectiveness of our method, we perform evaluation on SD1.5~\cite{rombach2022high}, SD2.1~\cite{sd2-1-base} and SDXL~\cite{podell2023sdxl}. We perform three unseen resolutions, with scaling factors of \(0.25\times0.25\), \(2\times2\), \(4\times4\) relative to the original training resolution. Specifically, we generate images of \(256\times256\), \(1024\times1024\), and \(2048\times2048\) for SD1.5 and SD2.1, while \(512\times512\), \(2048\times2048\) and \(4096\times4096\) for SDXL. We randomly select 1024 prompts to conduct evaluation from LAION-5B~\cite{schuhmann2022laion}, which contains 5 billion image-caption pairs.


\textbf{Evaluation metrics.} Following prior work, we report Frechet Inception Distance (FID)~\cite{heusel2017gans} and Kernel Inception Distance (KID)~\cite{binkowski2018demystifying} to evaluate the quality and diversity of generated images.
Following previous work~\cite{chai2022any,qiu2024freescale}, we use crop local patches to calculate the above metrics, defined as \(\text{FID}_{c}\) and \(\text{KID}_{c}\). Notably, for lower resolution generation on SD1.5 and SD2.1, the images cannot be further cropped, so they do not have \(\text{FID}_{c}\) and \(\text{KID}_{c}\). Additionally, we also report the CLIP score (Clip)~\cite{radford2021learning} and inference time (Time).

\subsection{Main Results.}
For lower resolution generation, we compare our method with SDXL~\cite{podell2023sdxl} Direct-Inference, Attn-SF~\cite{jin2023training} and ElasticDiffusion~\cite{haji2024elasticdiffusion}. For higher resolution, we compare with SDXL~\cite{podell2023sdxl} Direct-Inference, Attn-SF~\cite{jin2023training}, ScaleCrafter~\cite{he2023scalecrafter}, FouriScale~\cite{huang2024fouriscale}, HiDiffusion~\cite{zhang2023hidiffusion}, AccDiffusion~\cite{lin2024accdiffusion}, MegaFusion (without experimental setting at \(4096^2\) resolutions)~\cite{wu2025megafusion}, DiffuseHigh~\cite{kim2024diffusehigh}, FreCaS~\cite{zhang2024frecas}, DemoFusion~\cite{du2024demofusion} and FreeScale~\cite{qiu2024freescale}. Additionally, we integrate our method into ~\cite{he2023scalecrafter,huang2024fouriscale,qiu2024freescale}. \textbf{More results} on \textbf{higher-resolution}, \textbf{SD3}~\cite{esser2024scaling}, and comparisons with \textbf{Super-Resolution} are available in the supplementary materials.

\textbf{Quantitative results. }
In Table \ref{tab:main}, for lower resolution generation, our method achieves the best performance in almost metrics, demonstrating effectiveness in generating complete and detailed visual content. It’s important to note that the inference times for ElasticDiffusion are approximately \textbf{15} times longer than ours.
For higher resolution generation, our method further enhances the performance of three baselines, particularly in terms of \(\text{FID}_c\) and \(\text{KID}_c\). In the \(2\times2\) experiments on SD1.5 and SD2.1, ScaleCrafter+Our (integrated with our method) outperforms other approaches. In all experiments on SDXL, FreeScale+Our achieves the best scores on almost metrics.

\textbf{Qualitative results. }
In the Fig.~\ref{fig: compare_baseline}, we show the visual comparison results. For lower resolution, the generated results of our method have richer details and more completed structure compared with Attn-SF and ElasticDiffusion, including smaller scaling factor (\(0.25\times0.25\)), which demonstrates the powerful ability of our method. For higher-resolution generation, our method further reduces the small local repetition that appears in Scalecrafter and freescale on the 2x2 experiment. For the 4x4 experiment, the cat's eyes and ears in ScaleCrafter have obvious aliasing, and FreeScale fails to generate accessory on the chest. In comparison, our results have better visual quality.

\begin{figure}
    \centering
  \includegraphics[width=0.45 \textwidth]{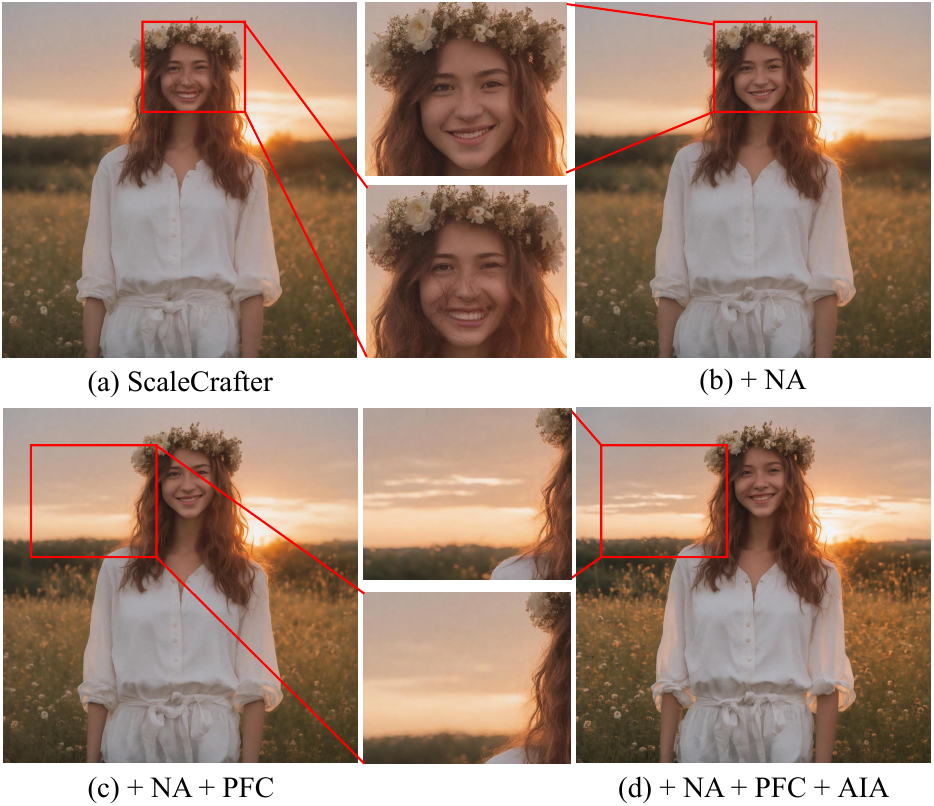}
  \\
  \caption{Qualitative results for ablation study.}
  \label{fig:ablation}
\end{figure}

\subsection{Ablation Study}
In this section, we use the SD2.1 and conduct a series of ablation experiments with 2x2 scale factor setting to verify the effectiveness of each component, as shown in Table~\ref{tab:Ablation studies}.

\textbf{Effect of Progressive Frequency Compensation ({PFC}).} As shown in Fig.~\ref{fig:ablation} (c). Although the dilated convolution significantly reduces the repetition issue, the background of generated images becomes blurred. Compared with (a), our PFC compensate the high-frequency information loss caused by the dilated convolution, making the background clearer and improving the global details.

\begin{table}[htbp]
  \centering
  \caption{Ablation studies for each Components in InfoScale}
    \begin{tabular}{ccccccc}
    \hline
    \textbf{PFC} &\textbf{AIA} & \textbf{NA} &\textbf{FID} & \textbf{KID} & \textbf{FID$_c$} & \textbf{KID$_c$}\\
    \hline
    \checkmark &  & & 63.99   & 0.005  &  37.40  & 0.006 \\
    \checkmark &  & \checkmark& 63.48   & 0.005  & 37.08  & 0.007 \\
    \checkmark & \checkmark & & 62.95   & 0.005  & 36.73  & 0.006 \\
    \checkmark & \checkmark &\checkmark & 62.73   & 0.004  & 36.47  & 0.006\\
    \hline
    \end{tabular}
  \label{tab:Ablation studies}%
\end{table}

\textbf{Effect of Adaptive Information Aggregation (AIA).} DSAttn has better information aggregation ability as shown in Fig.~\ref{fig:low_high_attn} . We further utilization AIA module in higher resolution to balance local and global information, as shown in Fig.~\ref{fig:ablation} (d). After adopting AIA, the image details are further improved. The comparison results of DSAttn and AIA can be found in the supplementary material.

\textbf{Effect of Noise Adaptation (NA).} 
We use NA module to suppress the distribution of information in the central region to mitigate the phenomenon of repetitive content in higher resolution. As shown in Fig.~\ref{fig:ablation} (b), the messy hair on the face is successfully removed after using NA module.
\section{CONCLUSION}
We propose \textbf{InfoScale}, an information-centric variable-scaled image generation framework, achieving effective information utilization for DMs. We believe that information amount of the generated image is different across resolutions, leading to the information conversion procedures need to be varied when converting the initial noise to variable-scaled images. We design Progressive Frequency Compensation module, Adaptive Information Aggregation module and Noise Adaptation module to address these challenges. Our method is plug-and-play for DMs and extensive experiments demonstrate the effectiveness of our method.

{
    \small
    \bibliographystyle{ieeenat_fullname}
    \bibliography{main}
}


\end{document}